\documentclass[review]{elsarticle}

\usepackage{lineno,hyperref}
\usepackage{graphicx}
\usepackage{amsmath,amssymb} % define this before the line numbering.

\usepackage{color}
\usepackage{xspace} 
\journal{Pattern Recognition}

\usepackage{hyperref}
\hypersetup{
	linkcolor=blue,          % color of internal links 
	citecolor=green,        % color of links to bibliography
}
\usepackage{algorithm}
\usepackage{algorithmic}
\usepackage{bbm}
\newcommand {\Forall} {\forall\,}

\newcommand{\U}{\mathbf{1}}
\renewcommand{\O}{\mathbf{0}}

\newcommand{\idc}[1]{\mathbbm{1}_{\{#1\}}}

\newcommand {\RR} {{\mathbb{R}}}
\newcommand{\R}{\mathcal{R}}

\newcommand {\PP} {{\mathbb P}}

\newcommand {\Ss} {{\mathcal S}}
\newcommand {\II} {{\mathcal I}}
\renewcommand {\P} {{\mathcal P}}
\newcommand {\M} {{\mathcal A}}
\renewcommand {\R} {{\mathcal R}}
\def \p {P}

\newcommand {\dotp}[2] {{\left\langle #1, #2 \right\rangle}}
\DeclareMathOperator {\proj} {\text{Proj}}

\DeclareMathOperator {\argmin} {argmin}
\DeclareMathOperator {\argmax} {argmax}
\DeclareMathOperator {\vect} {vec}

\newcommand {\Argmin}[1] {\underset{#1}{\argmin}}
\newcommand {\Argmax}[1] {\underset{#1}{\argmax}}
\newcommand {\norm}[1] {{\lVert{#1}\rVert}}

\def\etal{{\em et al}\@\xspace}
\def\ie{i.e.\@\xspace}
\def\eg{e.g.\@\xspace}

\usepackage{multirow}
\DeclareMathOperator{\tr}{tr}
\newcommand{\ind}[1]{\mathbbm{1}_{#1}}
\newcommand{\card}[1]{\lvert{#1}\rvert}
\newcommand{\Sig}{\mathrm{\Sigma}}
\newcommand*{\lcdot}{\raisebox{-0.6ex}{\scalebox{1.8}{$\cdot$}}}
\newcommand{\rR}{R\xspace}
\newcommand{\rP}{P\xspace}

\begin{document}

\begin{frontmatter}

\title{Automatic discovery of discriminative parts as a quadratic assignment problem}
%\tnotetext[mytitlenote]{Fully documented templates are available in the elsarticle package on \href{http://www.ctan.org/tex-archive/macros/latex/contrib/elsarticle}{CTAN}.}

\author{Ronan Sicre$^1$, Julien Rabin$^2$, Yannis Avrithis$^1$, Teddy Furon$^1$, Frederic Jurie$^2$}
\address{$^1$INRIA - Campus Beaulieu, Rennes, France\\
	$^2$GREYC UMR 6072 - University of Caen, France 
}
%% Group authors per affiliation:
%\author{Elsevier\fnref{myfootnote}}
%\address{Radarweg 29, Amsterdam}
%\fntext[myfootnote]{Since 1880.}
%
%%% or include affiliations in footnotes:
%\author[mymainaddress,mysecondaryaddress]{Elsevier Inc}
%\ead[url]{www.elsevier.com}
%
%\author[mysecondaryaddress]{Global Customer Service\corref{mycorrespondingauthor}}
%\cortext[mycorrespondingauthor]{Corresponding author}
%\ead{support@elsevier.com}
%
%\address[mymainaddress]{1600 John F Kennedy Boulevard, Philadelphia}
%\address[mysecondaryaddress]{360 Park Avenue South, New York}

\begin{abstract}
Part-based image classification consists in representing categories by small sets of discriminative parts upon which a representation of the images is built. This paper addresses the question of how to automatically learn such parts from a set of labeled training images. The training of parts is cast as a quadratic assignment problem in which optimal correspondences between image regions and parts are automatically learned. The paper analyses different assignment strategies and thoroughly evaluates them on two public datasets: {\em Willow actions} and {\em MIT 67 scenes}. State-of-the art results are obtained on these datasets.  
 % Several optimization methods feet such a problem and an evaluation is carried out on their performance.
% Interestingly our study shows that a soft assignments of the regions to the parts model offers overall better performances.
% Furthermore, we propose a fast and well performing method to extract region description, based on the convolutional layers of very deep CNN architectures, leading to high performance on challenging datasets.

%In this work, based on a set of constraint, we  several optimization methods

\end{abstract}


\end{frontmatter}

%\linenumbers

\section{Introduction}
\label{intro}

The representation of images as set of patches has a long history in computer vision, especially for object recognition~\cite{boureau2010learning}, image classification~\cite{doersch2012makes} or object detection~\cite{lim2013sketch}. Its biggest advantages are the robustness to spatial transformations (rotation, scale changes, etc.) and the ability to focus on the important information of the image while discarding clutter and background.

Part-based recognition raises the questions of i) how to automatically identify what are the parts to be included in the model and ii) how to use them to take a decision \eg to assign a category to an image. As an illustration, \cite{Ullman2001gh} proposed to select informative patches using an entropy based criterion while the decision relies on a naive-Bayes classifier.
%where the class-conditional probabilities are  expressed as the product of individual parts probabilities.
Following \cite{Ullman2001gh}, recent approaches separate the construction of the model (\ie the learning of the parts) and the decision function ~\cite{Juneja13} ~\cite{doersch13}. The reason behind this choice is that the number of candidate regions in the training images is very large and would lead to a highly non-convex decision function.

Optimizing both (parts and decision) is however possible for simple enough part detectors and decision functions. For instance, \cite{parizi2014automatic} unifies the two stages by jointly learning the image classifiers and a set of shared parts. %They first generate a set of initial parts by randomly sampling candidates and selecting a good subset.
Their claim is that the definition of the parts is directly related to the final classification function.  

While this argument is true, the objective function of this joint optimization is highly non-convex with no guaranty of convergence. We believe that deciding which one of the two alternatives -- the joint optimization vs separate one -- is still an open problem. As an insight, the two stage part-based model of~\cite{Sicre15A} performs better than the joint learning of~\cite{parizi2014automatic}.
%\color{blue}
We note that there are other differences between the two approaches, \eg ~\cite{parizi2014automatic} models both positive and negative parts while~\cite{Sicre15A} focuses only on the positive ones.
%\normalcolor

Interestingly, \cite{Sicre15A} addresses the learning of parts as an assignment problem. On one hand, regions are sampled randomly from the training images. On the other hand, the model is considered as a set of parts. The assignment is constrained by imposing that each part should be assigned to one image region in each positive image (those belonging to the category to be modeled). This results in a bipartite graph linking parts and regions. 

The assignment problem of~\cite{Sicre15A} poses the learning of part-based models in a very appealing way, yet their solution is based on heuristics leaving room for improvements. 
%\color{blue}
%They observed that stopping the optimization before convergence leads to better performance. This raises the question of whether the constraints should be binary or not.
%\normalcolor
%\color{blue} 
This paper's contribution is an extensive study of this assignment problem: We first present of a well-founded formulation of the problem and propose different solutions in a rigorous way.
These different methods are evaluated and compared on two different datasets and state-of-the-art performance is obtained. 
%\color{red}
By experimenting with improvements in the underlying description and encoding, we demonstrate that the benefit of our part learning methodology remains complementary to the benefit of more powerful visual representations obtained by state of the art deep learning approaches.
%\normalcolor

%This paper makes the following contributions:
%\begin{itemize}
%\item First, The paper gives a well-founded formulation of the assignment problem and proposes different solutions in a rigorous way.
%\item Secondly, several improvements are proposed for region description, part encoding, and region sampling.
%\end{itemize}

%The paper is organized as follows: Section 2 gives the related works, Section 3 presents the formulation of the problem while Section 4 introduces the optimization frameworks. Finally, Section 5 is devoted the experimental validation. 

\section{Previous work}

%This paper investigates the use of parts in image classification tasks.
Image classification has received a lot of attention during the last decades, with most of the related approaches focused on models based on aggregated features \cite{CsurkaSLCV2004,PerroninECCV2010} or the Spatial Pyramid Matching \cite{LazebnikCVPR2006}. This was before the Convolutional Network revolution \cite{krizhevsky12} still at the heart of most of the recent methods \cite{Simonyan14c}.

Several authors have investigated part-based models in which some parts of the image are combined in order to determine if a given object is depicted. This is in contrast to aggregation approaches where all the image regions are pooled without selecting sparse discriminative parts. For instance,~\cite{singh12} discovers sets of regions used as mid-level visual representation; the regions are selected for being representative (occurring frequently enough) and  discriminative (different enough from others), during an iterative procedure which alternates between clustering and training classifiers. Similarly, \cite{Juneja13}  addresses this problem by learning parts incrementally, starting from a single part occurrence with an Exemplar SVM and collecting more occurrences from the training images.

In a different way, \cite{doersch13} poses the discovery of visual elements as a discriminative mode seeking problem solved with the mean-shift algorithm: it discovers visually-coherent patch clusters that are maximally discriminative. In \cite{Maji2013}, Maji \etal investigate the problem of parts discovery when some correspondences between instances of a category are known. 
%\color{red}
The work of~\cite{SuPo13} bears several similarities to our work in the encoding and classification pipeline. However, parts are assigned to regions using spatial max pooling without any constraint on the number of regions a part is assigned to (from zero to multiple); given this fixed assignment, part detectors are optimized using stochastic gradient descent.
%\normalcolor

The recent papers related to part-based models are those of Sicre \etal~\cite{Sicre15A} and Parizi \etal~\cite{parizi2014automatic}. As said before, the part-based representation of~\cite{parizi2014automatic} relies on the joint learning of informative parts (using heuristics that promote distinctiveness and diversity) and linear classifiers trained on vectors of part responses. On the other hand, Sicre \etal \cite{Sicre15A} follow the two stage formulation, formulating the discovery of parts as an assignment problem. We also mention the recent and unpublished work of Mettes \etal \cite{Mettes15} arguing that image categories may share parts and proposing a method to model them as such. 

Finally, this paper is related to the assignment problem which finds a maximum weight matching in a weighted bipartite graph. A survey on this topic is the work of Burkard~\etal~\cite{Burkard_Assignment}.

\section{Discovering and Learning Parts}
\label{sec:DiscoLearn}

The studied % Our
approach comprises three steps:
(i) distinctive parts are discovered and learned for every category;
(ii) a global image signature is computed based on the presence of these parts; and
(iii) image signatures are classified by a linear SVM.
This paper focuses on the first step. For each category, we learn a set of $\p$ distinctive parts which are representative and discriminative.

This section presents different ways to formalize this task giving birth to interesting optimization alternatives in Sect.~\ref{sec:optim}.
We first present the parts learning problem as defined in~\cite{Sicre15A}.
We show that it boils down to a concave minimization under non convex constraints, which is recast as a quadratic assignment problem.
%We further show how to solve such a problem approximately using convex relaxation.
%Next, initialization and encoding are presented.
%Finally, we present our improved region description.

\subsection{Notation}
$X^\top$ and $\tr(X)$ are the transpose and trace of matrix $X$; $\vect(X)$ is the column vector containing all elements of $X$ in column-wise order. Given matrices $X, Y$ of the same size, $\dotp{X}{Y} = \sum_{i,j} X_{ij} Y_{ij}$ is their (Frobenius) inner product, $\norm{X}$ and $\norm{X}_F = \sqrt{\dotp{X}{X}}$ are the spectral and Frobenius norms. The Euclidean norm of vector $x$ is $\norm{x} = \sqrt{\dotp{x}{x}}$. Vector $x_{i\lcdot}^\top$ ($x_{\lcdot j}$) denotes the $i$-th row (resp. $j$-th column) of matrix $X$. The $n \times n$ identity matrix is denoted as $I_n$, while vector $\U_n$ (matrix $\U_{m \times n}$) is an $n \times 1$ vector (resp. $m \times n$ matrix) of ones. $\ind{\M}$ is the indicator function of set $\M$ and $\proj_{\M}$ is the Euclidean projector onto $\M$.

Following~\cite{Sicre15A}, we denote by $\II^+$ with $n^+=\card{\II^+}$ the set of images of the category to be modeled, \ie positive images, while $\II^-$ represents the negative images. The training set is $\II=\II^+\cup \II^-$ and contains $n=\card{\II}$ images.
A set of regions ${\R}_I$ is extracted from each image $I\in\II$. The number of regions per image is fixed and denoted $\card{\R}$. The total number of regions is thus $R = n\card{\R}$. $\R^+$ is the set of regions from positive images whose size is $R^+ = n^+\card{\R}$.

Each region $r \in {\cal R}_I$ is represented by a descriptor $x_r \in \RR^d$. In this work, this descriptor is obtained by a CNN, and in particular it is the output of a convolutional or fully connected layer. More details are given in Section~\ref{sec:desc}. By $X$ ($X^+$) we denote the $d\times R$ (resp. $d\times R^+$) matrix whose columns are the descriptors of the complete training set (resp. positive images only).

\subsection{Problem setting}
A category is modeled by a set of parts $\P$ with $\card{\P}=\p$. We introduce the $\p \times R^+$ matching matrix $M$ associating image regions of positive images to parts. Element $m_{pr}$ of $M$ corresponds to region $r$ and part $p$. Ideally, $m_{pr}=1$ if region $r$ represents part $p$, and $0$ otherwise. By $M_I$ we denote the $\p \times \card{\R}$ submatrix of $M$ that contains columns $r\in\R_I$ corresponding to image $I$.

We keep the requirements of~\cite{Sicre15A}: (i) the $\p$ parts are different from one another, (ii) each part is present in every positive image, (iii)  parts should occur more frequently in positive images than in negative ones. The first two requirements define constraints on the admissible set $\M$ of $M$:
% \begin{equation}\label{eq_Mjk}
% 	\M \triangleq \left\{M \in  \{0,1\}^{P\times R}, 
% 	\forall\, I \in {\II}^+,\, 
% 	\begin{array}{ll}
% 	\forall\, r \in {\R}_I \, 
% 	    & \sum_{p \in {\cal P}} M(p,r) \leq 1,
% 	\\
% 	\Forall p \in {\P} 
% 	    & \sum_{r \in {\R}_I} M(p,r) = 1 
% 	\end{array}
% 	\right\}
% 	.
% \end{equation}
\begin{equation}\label{eq_Mjk}
	\M \triangleq \left\{
	    M \in \{0,1\}^{\p\times R^+}:
	    M^\top \U_\p \le \U_{R^+} \text{ and }
	    M_I \U_{\card{\R}} = \U_\p \text{ for } I\in\II
	\right\}.
\end{equation}
This implies that each sub-matrix $M_I$ is a \emph{partial assignment} matrix.
Observe that the set $\M$ is not convex.
The third assumption is enforced by Linear Discriminant Analysis (LDA): given matching matrix $M$, the model $w_p(M)$ of part $p$ is defined as
% \begin{equation}\label{eq_lda}
%  w_M(p) \triangleq 
% \Sigma^{-1}\left(
% \frac{\sum_{r\in \R_I,\, I\in\II^+} \, M(p,r) \, x_r }{\sum_{r\in \R_I,\, I\in\II^+} M(p,r)}
% - \bar{x}\right)\; 
% %\frac{\sum_{s\in \R_J,\, J\in\II} \, x_s}{R}\right)\;
% \in \RR^d,
% \end{equation}
\begin{equation}\label{eq_lda}
    w_p(M) \triangleq \Sig^{-1}\left(
        \frac{\sum_{r\in\R^+} m_{pr} x_r }{\sum_{r\in\R^+} m_{pr}} - \mu
    \right) 
    = \Sig^{-1}\left( \frac{1}{n^+} X^+ m_{p\lcdot}^\top - \mu \right),
\end{equation}
where $\mu = \frac{1}{n} X\U_R$ and $\Sig = \frac{1}{n} (X- \mu\U_{R}^\top)(X- \mu\U_{R}^\top)^\top$ are the empirical mean and covariance matrix of region descriptors over all training images. The similarity between region $r$ and a part $p$ is then computed as the inner product $\dotp{w_p(M)}{x_r}$.

For a given category, we are looking for an optimal matching matrix
\begin{eqnarray}
    &M^\star \in \arg\max_{M\in\M} J(M)\label{eq:problem}\\
    &J(M) \triangleq \sum_{p\in {\P}} \sum_{r\in \R^+} m_{pr} \, \dotp{w_p(M)}{x_r}
    = \dotp{M}{W(M)^\top X^+} \,,\label{eq:objective}
\end{eqnarray}
%that maximizes the objective function
% \begin{equation}\label{eq:problem}%\label{objective}
%  {M}^\star =  \underset{M \in \M} \argmax  \;\, J(M) \triangleq \sum_{p \in {\cal P}} \sum_{I\in {\II^+}} \sum_{r\in \R_I} M(p,r) \, (w_M(p)^\top x_r)
%  \;\,.
% \end{equation}
%\begin{equation}\label{eq:objective}\end{equation}
where $W(M)$ is the $d \times \p$ matrix whose columns are $w_p(M)$ for all parts $p\in \P$.

%-------------------------------------------------------------

\subsection{Recasting as a quadratic assignment problem}

The previous formulation limits optimization to alternatively resorting to~\eqref{eq_lda} and~\eqref{eq:objective}, as done in~\cite{Sicre15A}.
%implies that given $M$ one can compute the part models $W = (w_p)_{p\in\P}$ by~\eqref{eq_lda}, while fixing the part models $W$ one can update the matching matrix $M$ by maximizing $J(M)$ in~\eqref{eq:objective}. This limits optimization to a particular alternating scheme as in~\cite{Sicre15A}.
Here, we express $J$ as a function of $M$ without $W$, recasting~\eqref{eq:problem} as a quadratic assignment problem and opening the way to a number of alternative optimization algorithms.
We define similarity matrix $C(M) \triangleq {W(M)^\top X^+}$.
Its entries represent the similarities between parts and regions. 
According to LDA~\eqref{eq_lda}, $W(M) = \Sig^{-1}\left( \frac{1}{n^+} X^+ M^\top - \mu \U_\p^\top \right)$, which in turn gives
\begin{equation}\label{eq:sim}
    C(M) = \left( \frac{1}{n^+} M {X^+}^\top - \U_\p \mu^\top \right) \Sig^{-1} X^+
    = MA - B,
\end{equation}
where $n^+ \times n^+$ matrix $A = \frac{1}{n^+} {X^+}^\top \Sig^{-1} X^+$
is symmetric and positive definite % (because we can write $A = S^\top S$ since $\Si$ is symmetric and positive).
and $\p \times n^+$ matrix
$B = \U_\p \mu^\top \Sig^{-1} X^+$
has identical rows (rank 1). 
Now, observe that problem \eqref{eq:problem} is equivalent to
\begin{eqnarray}
   & \text{find } \quad M^\star \in \arg\min_{M\in \M} \; J_0(M)\label{eq:main_problem}\\
&J_0(M) \triangleq \dotp{M}{B - MA}=\vect(M)^\top Q \vect(M) + \vect(B)^\top \vect(M)    \label{eq:main_objective}
\end{eqnarray}%
%where
%\begin{eqnarray}\label{eq:main_objective}
%    \text{where } \quad J_0(M) &\triangleq&  -J(M) =\dotp{M}{B - MA}\nonumber
%    \\
%    &=& \vect(M)^\top Q \vect(M) + \vect(B)^\top \vect(M)
%\end{eqnarray}
for a $PR^+ \times PR^+$ matrix $Q$ that is a function of $A$ only.
%for $Q$ a matrix function of $A$ only.
This shows that our task is closely related to the \emph{quadratic assignment problem}~\cite{Burkard_Assignment}, a NP-hard combinatorial problem. 
Moreover, in our setting, 
%both the constraint set and the objective are non-convex. 
the objective function to be minimized is strictly concave.

%Equivalently, we have
%\begin{equation}\label{eq:main_min_problem}
%\Argmin{M \in \M} \;  \left\{J_0(M) := \dotp{M}{C(M)} =  \dotp{M}{B-MA} \right\}%= \sum_{p,r} M_{p,r} {C(M)}_{p,r}
%\end{equation}
\medskip
This new formalism enables to leverage a classical procedure in optimization: the convex relaxation.

%We consider two approaches from the literature:
%\begin{enumerate}
%\item transforming the concave minimization problem into a convex minimization thanks to %quadratic 
%regularization;
%\item while relaxing of the non-convex binary constraints into linear constraints.
%\end{enumerate}
% We consider in the next paragraphs two different approaches from the literature.

%-------------------------------------------------------------
\subsection{Convex relaxation with entropic regularization}\label{sub:SoftAssign}
%  (soft linear assignment)

%We begin with the linear assignment problem obtained for \textbf{fixed cost matrix} $C$.

In the specific case of \textbf{fixed cost matrix} $C$, the previous problem becomes tractable.

%However, recall that solving a linear assignment problem (\ie bi-partite graph matching) is by itself numerically demanding, with a complexity about $O(N^3)$ ($N$ being the size of the data). To decrease the computational complexity of such a strategy, each linear assignment problem is approximated using entropic regularization (\ie into a soft assignment problem described in \ref{sec:SoftAssign}) which is solved by the fast Sinkhorn algorithm.

\subsubsection{Convex relaxation}
Solving a linear assignment problem
%(\ie bipartite graph matching) is by itself
is numerically demanding, with a complexity about $O(N^3)$~\cite{Burkard_Assignment}.
%($N$ being the size of the graph), but
It can be done exactly with dedicated methods, such as the Hungarian algorithm; or equivalently, with linear programming methods that assume convex relaxation of the binary constraints, \ie considering bi-stochastic instead of permutation matrices.

\subsubsection{Soft assignment}
To reduce the complexity, the problem is approximated using negative-entropy regularization.
%(\ie into a soft assignment problem described in \ref{sec:SoftAssign}) which can be solved by the fast Sinkhorn algorithm.
Considering a $\p \times \p$ bi-stochastic matrix $M$, 
the soft-assignment problem is 
% obtained when considering a negative-entropy regularization
\begin{equation}\label{eq:soft-assign}
    \Argmin{\substack{M\in \R_+^{\p \times \p},\\ M \U_{\p} = \U_\p, M^\top \U_\p = \U_{\p}}}
    %E_\beta = 
    \left\{ \dotp{C}{M} - \frac1\beta E(M) = \dotp{C}{M} + \frac1\beta \dotp{\log(M)}{M}
    \right\}
\end{equation}
where $E$ is the entropy of the bistochastic matrix $M$, and $\beta >0$ is the regularization parameter.
As $\beta$ increases, the problem converges to the hard-assignment problem. Paper~\cite{Sicre15A} uses the Sinkhorn algorithm~\cite{Sinkhorn67}, which normalizes iteratively the rows and the columns of $M$ to one, initializing from the regularized cost matrix $\exp(\beta C)$.
%\red{mettre algo de sinkhorn ?}

Observe that in our setting $M$ is not square as we consider partial assignments between $\p$ rows and $R>\p$ columns. To solve this more general problem, a simple trick is to add as many rows than required and to define a maximal cost value when affecting columns to them.

Soft assignment has gained a lot attention because it solves large scale problems~\cite{Peyre_SIGGRAPH_15}.
However, a major limitation is the loss of sparsity of the solution.
As a consequence, approximate solutions of the linear soft-assignment are not suitable for our problem, as observed in experiments. 
%This has been confirmed experimentally.
We describe in the section~\ref{sec:Iterated-SoftAssign} how the authors of \cite{Sicre15A} have circumvented this problem by iterating soft assignment.

%-------------------------------------------------------------
\subsection{Convex relaxation with quadratic regularization}
\label{sub:relax}

%Framing the part learning task into a quadratic assignment problem~\eqref{eq:main_problem} enables to leverage a classical procedure in optimization: the convex relaxation. It consists in
%\begin{enumerate}
%\item transforming the concave minimization problem into a convex minimization thanks to quadratic regularization;
%\item while relaxing of the non-convex binary constraints into linear constraints.
%\end{enumerate}

\begin{figure}[tb]
\centering
	\includegraphics[width=0.31\linewidth]{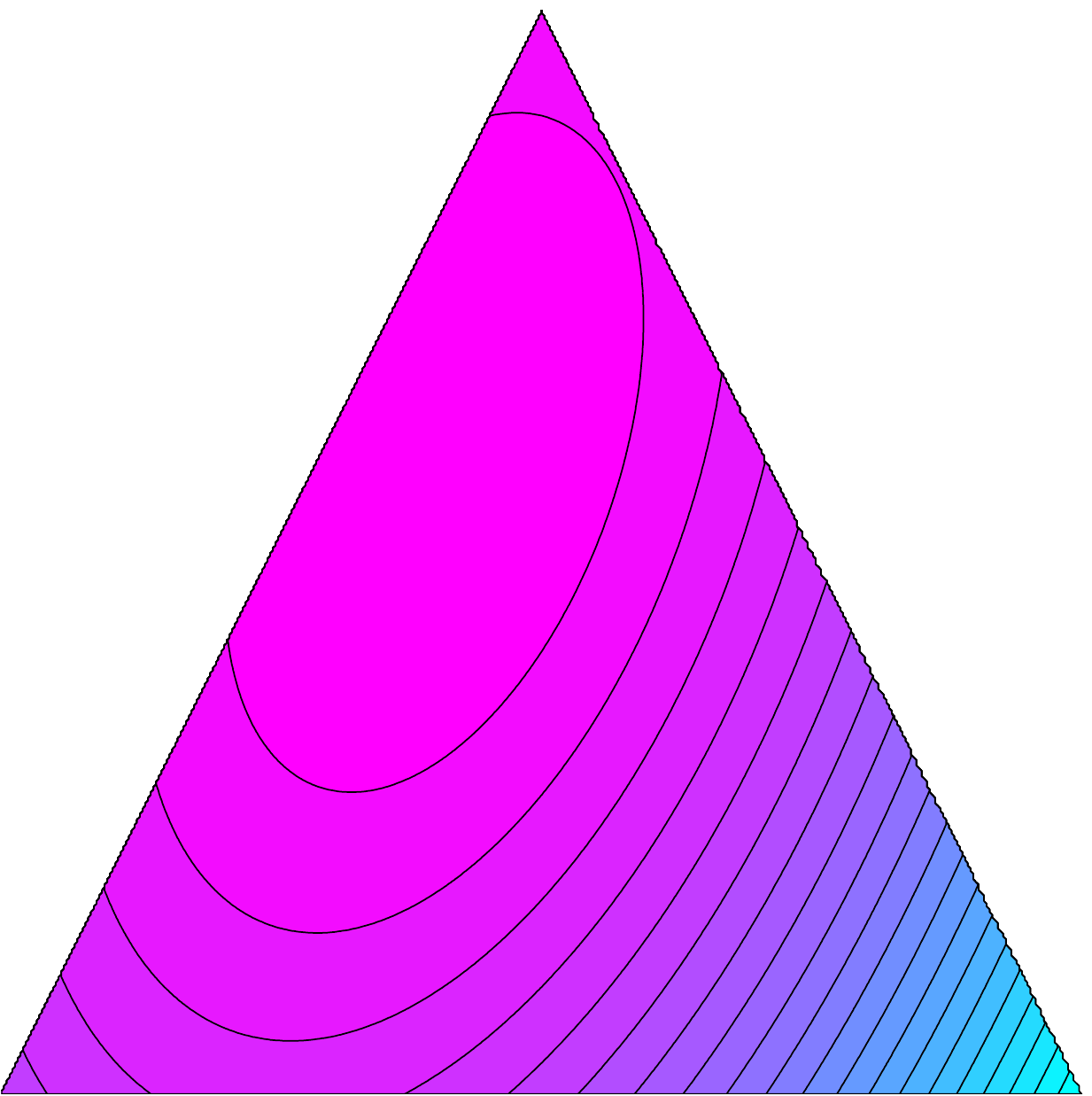}
	\includegraphics[width=0.31\linewidth]{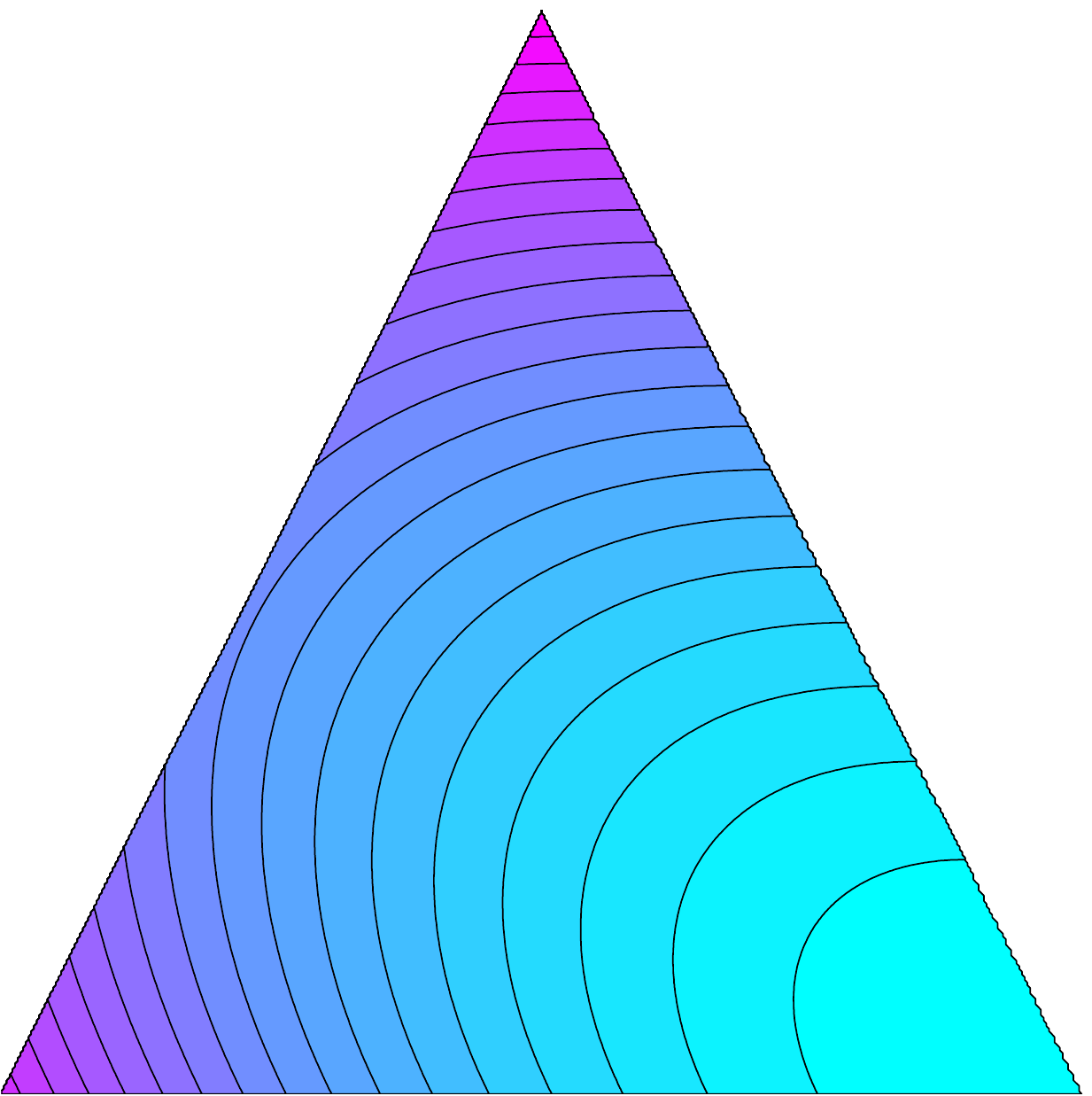}
	\includegraphics[width=0.31\linewidth]{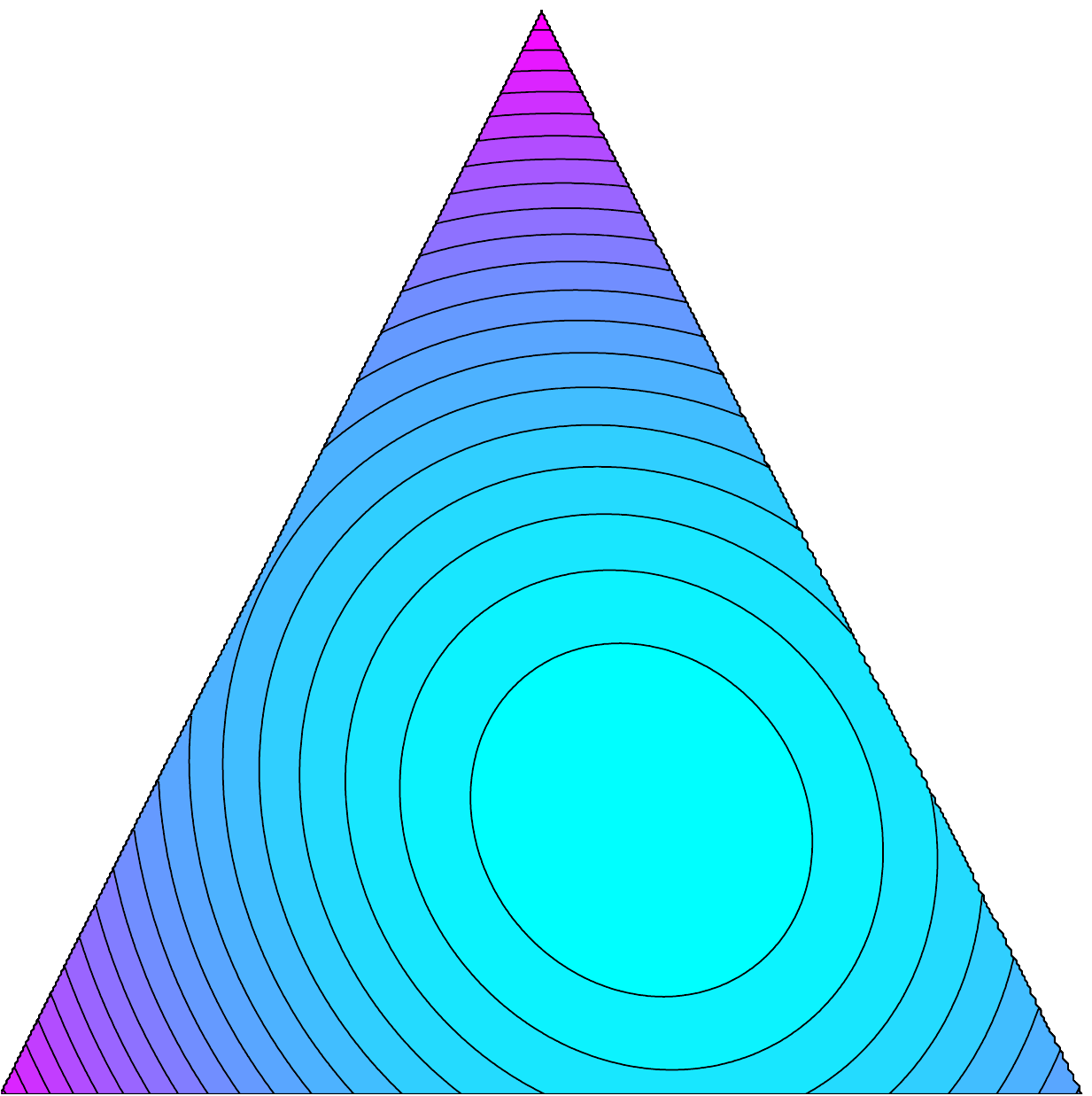}
\caption{\textbf{Illustration of the convex relaxation of our assignment problem in 3D.}
\it
Black lines are level-sets of the objective function $J_0$ in the plane of the simplex, which is a triangle in $\RR^3$. Lower values are displayed in cyan, larger in magenta.
(Left) The original problem is the minimization of a concave quadratic function over matching matrices, that lies on the vertices of the simplex.% of bi-stochastic matrices. 
(Middle) A small quadratic regularization of the objective function together with the relaxation of the constraint
%(optimization over the simplex)
preserves the solution.% (location of the global minimum).
(Right) A too large regularization yet shifts the minimum inside the simplex%, instead of on the vertices
, thus giving less sparse solutions.} 
\label{fig:illustration_relaxation}
\end{figure}

We consider now the quadratic regularization of the problem (see Figure~\ref{fig:illustration_relaxation}):
% of the resulting quadratic convex relaxation
%The new objective function is
\begin{equation}
\begin{split}
J_\rho(M) 
    	& \triangleq \dotp{M}{M(\rho I_{n^+}-A) + B}
    	%= \langle \rho M-AM+B,M \rangle 
    	%= \langle A_\rho M+B,M \rangle
    	= \dotp{M}{B-MA} + \rho \lVert M \rVert_F^2
    	\\
    	& =  J_0(M) + \rho \, \p \, n^+ \text{ for } M \in \M.
    	\\
\end{split}
\end{equation}
This means that provided $M\in\M$, $J_\rho(M)$ and $J_0(M)$ differ by a constant.
Therefore, the minimizers of $J_\rho$ on $\M$ are the minimizers of $J_0$, for any value of $\rho$.
Indeed, if $\rho$ is sufficiently large ($\rho > \norm{A}$), $J_\rho$ becomes convex (see Figure~\ref{fig:illustration_relaxation}).

We also relax the constraints:
%, in order to enable soft-assignment of several regions for each part, we consider the relaxed problem
%\begin{equation}
%\Ss \triangleq \left\{ M \in [0,1]^{\p\times R^+}:\;   M^\top \U_\p \le \U_{R^+},\mbox{ and } M_I \U_{|\R|} = \U_\p \text{ for } I \in \II^+ \right\}. \label{eq:relaxed_constraint}
%\end{equation}
\begin{subequations}
 \label{eq:relaxed_problem}
 \begin{align}
    & M^\star = \arg\min_{M\in \Ss} \; J_\rho(M)
 \label{eq:relaxed_objective}
    \\
    & % simplex definition
	%\Ss = \{M \in \R_+^{P\times R}, \U_P^\top M \le \U_R^\top, M\U_R = %\U_P\}
	%\Ss = \{M \in [0,1]^{P\times R},\;\Forall I \in \II^+,\; \Forall r %\in \R(I) \sum_{p \in \P} M(p,r) \le 1, \Forall p \in \P \, \sum_{r \in \R(I)} M(p,r) = 1\}
	\Ss \triangleq \left\{ M \in [0,1]^{\p\times R^+}:\;   M^\top \U_\p \le \U_{R^+},\mbox{ and } M_I \U_{|\R|} = \U_\p \text{ for } I \in \II^+ \right\}. \label{eq:relaxed_constraint}
\end{align}
\end{subequations}
%Recall that $M_I$ refers to the columns of $M$ related to image $I$.
In words, domain $\Ss$ is the convex hull of the set $\M$ and we will refer it to as a simplex.
Yet, in general, $J_\rho(M) \not= J_0(M) + \rho \p n^+$ when $M \in \Ss\setminus\M$.
We may find different solutions than problem~\eqref{eq:main_problem}, as illustrated in Figure~\ref{fig:illustration_relaxation}.
Over-relaxing the problem for the sake of convexity is not interesting as it promotes parts described by many regions instead of a few ones. 
%(less sparse solutions). 
Indeed, when $\rho > \norm{A}$, the minimum of $J_\rho$ is achieved for the rank 1 matrix $\tfrac{1}{2}B A_\rho^{-1}$ which may lie inside $\Ss$.
This shows that $\rho$ implicitly controls the number of regions used to describe parts.

%\red{analyse trop precoce a deplacer ?}

\section{Optimization}\label{sec:optim}

The previous section formalizes the part learning task into optimization problems.
This section now presents various methods to numerically solve them.
We see two kinds of techniques: i) hard assignment optimization directly finding $M^\star\in\M$, ii) soft assignment optimization (Sect.~\ref{sub:SoftAssign} and~\ref{sub:relax}) that finds $M^\star\in\Ss$. 

This latter strategy is not solving the initial problem. However, as already observed in~\cite{Sicre15A} and~\cite{SoftAssign_VS_MaxPooling} for classification, soft-assignment affecting several regions to describe a part, may provide better results. This lesson learned from previous works deserves an experimental investigation in our context.

\subsection{Hard assignment methods}
\label{sec:learning}

\subsubsection{Hungarian Algorithm}\label{sec:Hungarian}

As mentioned in Sect.~\ref{sub:SoftAssign}, problem \eqref{eq:problem}, when the cost matrix $C(M)$ is fixed, is a variant of the \emph{linear assignment problem} for which several dedicated methods give an exact solution.
%, such as 
%interior point methods, simplex algorithm the auction algorithm or the Hungarian algorithm. 
Solving this approximated problem can be seen as computing the orthogonal projection of the matrix $C(M_0)$
($M_0$ being an initial guess, see Section~\ref{sec:init})
onto the set $\M$
\begin{equation}
    M^\star_{hun} = \proj_{\M} \left( C(M_0) \right) = \Argmax{M \in \M} \; {\dotp{M}{C(M_0)}}.
\end{equation}
% ATTENTION : il ne faut pas enlever M_0, ou alors changer le nom de variable dans l'argmax 
%In our setting, it involves for each image $I \in \II^+$ a projection of the corresponding sub-matrix $C(M_0)_I$ onto the set of partial assignment matrices. To compute efficiently such a projection,
In our setting, we use the fast Hungarian algorithm variant of \cite{bougleux_edition_preprint}.
The experimental section shows that this gives surprisingly good results in comparison to more sophisticated methods.

\subsubsection{IPFP}\label{sec:IPFP}
The Integer Projected Fixed Point (IPFP) method~\cite{IPFP} can be seen as the iteration of the previous method, alternating between similarity matrix $C(M)$ update and projections onto the constraints set $\M$.
More precisely, a first order Taylor approximation of the objective function is maximized (\eg using the Hungarian algorithm) and combined with a linesearch (see Algorithm~\ref{algo:IPFP}).
This approach guaranties the convergence to a local minimizer of $J(M)$ on the set $\M$.

\begin{algorithm}
\caption{IPFP algorithm for problem \eqref{eq:main_problem}}
\begin{algorithmic}\label{algo:IPFP}
\STATE \textbf{Init:} $M_0$, \textbf{set:} $k\leftarrow 0$, $M_{-1} \leftarrow \O$
\WHILE{$M_{k+1} \not= M_{k}$}
    \STATE $k \leftarrow k+1$
	\STATE $G_k \leftarrow 2M_k A-B \quad \text{ (gradient $ \nabla J(M_k)$) }$
	\STATE $P_{k+1}  \leftarrow  \proj_{\M} \left(  G_k \right)\quad \text{ (projection using partial Hungarian algorithm~\cite{bougleux_edition_preprint}) }$
	\STATE $\Delta_{k+1}  \leftarrow  P_{k+1} - M_k$ %  \quad \text{ (approximated gradient) }
	\STATE $c_k \leftarrow \dotp{G_k}{\Delta_{k+1} }$ % 		\quad (c_k \ge 0)
	\STATE $d_k \leftarrow \dotp{\Delta_{k+1} A}{\Delta_{k+1} }$
	\STATE $t_k = \min(-\frac{c_k}{2d_k}, 1) \text{ if } d_k<0$ and $t_k = 1$ otherwise 
	%$$t_k = \begin{cases}
	%	1 & \text{ if } d_k > 0 \\
	%	\min(-\frac{c_k}{2d_k}, 1) & \text{ if } d_k \le 0
	%	\end{cases}
	%	\quad (t_k \in [0,1])
	%	\quad (linesearch) $$
	\STATE $M_{k+1} \leftarrow  t_k P_{k+1} + (1-t_k) M_{k}$ (linesearch)
\ENDWHILE
\STATE \textbf{Output:} $P_k$
\end{algorithmic}
\end{algorithm}

We observed that IPFP converges very fast
%, after a few iterations ($k\le 4$),
nevertheless without improving much results.
%from the direct projection discussed before.
This is explained by the very specific structure of our problem, where the quadratic matrix $Q$ of~\eqref{eq:main_objective} is very sparse and negative definite. 

%Note that a variant of IPFP studied in~\cite{???} consists in solving a sequence of problems starting from a relaxed convex objective to the non-convex target objective. \red{A FINIR} This method has not been tested in this paper. \red{I would remove this comment}

\subsection{Iterative Soft-assignment (ISA)}
\label{sec:Iterated-SoftAssign}

The strategy of~\cite{Sicre15A} referred here to as \emph{Iterative Soft-Assign} (ISA) solves a sequence of approximated linear assignment problems.
It is based on the rationale: if we better detect regions matching a part, we will better learn that part; if we better learn a part, we will better detect region matching that part. 
Hence, the approach iteratively assigns regions to parts by yielding a $M$ for a given $C(M)$ (Sect.~\ref{sub:SoftAssign}) and learns the parts by yielding $W(M)$ for a given $M$ thanks to LDA.
The assignation resorted to a soft-assign algorithm (see~\cite{SoftAssign} for instance) which is also an iterative algorithm solving a sequence of entropic-regularized problems  (Section~\ref{sub:SoftAssign}) that converges to the target one. 
The general scheme of the algorithm is drawn in Algorithm~\ref{algo:Iterated-SoftAssign}.
%This procedure is time consuming because one has to progressively reduce the regularization parameter to enforce an "hard assignment" matrix $M \in \M$.

%
%As previously discussed in Section~\ref{sec:SoftAssign}, solving a linear assignment problem (\ie bi-partite graph matching) is by itself numerically demanding, with a complexity about $O(N^3)$ ($N$ being the size of the data).
%To decrease the computational complexity of such a strategy, 
%each linear assignment problem is approximated using entropic regularization (\ie into a soft assignment problem described in ) which is solved by the fast Sinkhorn algorithm.

%More precisely, authors of \cite{Sicre15A} use a variant of the SoftAssign Quadratic Assignment algorithm \cite{SoftAssign} that converge to a local minimum of hard assignment problem under some conditions. The general scheme of the algorithm is drawn in \eqref{algo:Iterated-SoftAssign}.

\begin{algorithm}
\caption{Iterated-SoftAssign algorithm}
\begin{algorithmic}\label{algo:Iterated-SoftAssign}
\STATE \textbf{Init:} $M = M_0$
\WHILE{$M \not\in \M$}
    \STATE $\beta \leftarrow \beta \times \beta_r$ (decreases regularization)
	\WHILE{$M$ has not converged}
	    \STATE update $C(M)$ using definition \eqref{eq:sim}
	    \STATE update $M$ by solving linear Soft-Assignment problem~\eqref{eq:soft-assign}
    \ENDWHILE
\ENDWHILE
\end{algorithmic}
\end{algorithm}

The approach suffers from two major drawbacks: it is computationally demanding due to the three intricate optimization loops, and it is numerically very difficult to converge to an hard-assignment matrix (due to the entropy regularization).
Nevertheless, as reported in~\cite{Sicre15A}, the latter limitation turns out to be an advantage for this classification task. % as the method provides state-of-the-art performance.
Indeed, the authors found out that early stopping the algorithm actually improves the performance.
However, the obtained matrix $M$ does not satisfy the constraints (neither $\M$ nor $\Ss$). % and has to be manually

\subsection{Quadratic soft assignment with Generalized Forward Backward (GFB)} 

To address the relaxed problem~\eqref{eq:relaxed_constraint},
%{eq:relaxed_problem}, 
we split the
%rely on a splitting technique to obtain separable
constraints on the matching matrix $M$ for rows and columns:
for each row $m_{p\lcdot}$ and each column $m_{\lcdot r}$ of $M$
\begin{itemize}
    \item  $m_{p\lcdot} \in \PP \triangleq \{x \in \RR_+^{|\R|}: \, \dotp{x}{\U_{|\R|}} = 1\}$ is a vector summing up to 1; % \PP_P
    \item $m_{\lcdot r} %that sums at most to 1
    \in \PP_{\le} \triangleq \{x \in \RR^{\p}: \, \dotp{x}{\U_{\p}} \le 1\}$ is a vector that sums at most to $1$; % \PP_{|\R|}
\end{itemize}
%where non-negativity of columns is ensured by constraints on rows. 
%
Problem~\eqref{eq:relaxed_constraint} is then equivalent to the following
%convex optimization tools.
%we recast it as a quadratic minimization problem with independant constraints
%
\begin{equation}\label{eq:GFB_objective}
%\begin{split}
	\Argmin{M = M_1 = M_2 \,\in\, \RR^{\p\times R^+}} 
	J_\rho(M) + G_1(M_1) + G_2(M_2)
%\end{split}
\end{equation}
where
$G_1$ and $G_2$ respectively encode constraints on parts %(\emph{i.e.} rows that are simply referred to as $M_p$ for sake of clarity)
and regions:
%(\emph{i.e.} columns referred to as $M_r$)
$$
\begin{cases}  
    G_1(M) & = \sum_{p \in \P} \idc{m_{p\lcdot} \in \PP}  
    \\
    G_2(M) & = \sum_{I \in \II^+, r \in \R(I)} \idc{m_{\lcdot r} \in \PP_\le}  
\end{cases}
.
$$
%To address this new problem, we make use of
The General Forward Backward (GFB) algorithm~\cite{raguet-fadili-GFB} alternates between explicit gradient descent on the primal problem and implicit gradient ascent on the dual problem. It offers theoretical convergence guaranties in the convex case.
%The algorithm~\ref{algo:GFB} reads as follows:
%\begin{equation}\label{algo:GFB}
%\begin{cases}
%	& \nabla J_\rho(M) = 2 M A_\rho + B \quad \text{ (gradient) }\\
%	\forall \text{ (row) }p , & 
%		M^1_p  \leftarrow  M^1_p - M_p +  \, \proj_{\PP} \left( % 2M_p - M^1_p - \frac{1}{L} G_p \right)\\
	%\quad \text{\small ( projected gradient descent on simplex $\PP$) }\\
%	\forall I \in \II ,  \forall \text{ (column) } r, & 
%		 M^2_r  \leftarrow  M^2_r - M_r +  \, \proj_{\PP_\le} \left(  2M_r - M^2_r - \frac{1}{L} G_r \right)\\
%	& M \leftarrow \frac{1}{2} (M_1 + M_2)\\
%\end{cases}
%\end{equation}
%
%\scalebox{0.6}{
\begin{algorithm}
\caption{GFB algorithm for problem \eqref{eq:GFB_objective}}
\label{algo:GFB}
\small
\begin{algorithmic}
\STATE $M\gets M_0$ \quad \text{ (initialization) }
\WHILE{not converge} 
        \STATE $\nabla J_\rho(M) = 2 M A_\rho + B$ \quad \text{ (gradient) }
        \STATE update $M_1$: $m^1_{p\lcdot}  \leftarrow  m^1_{p\lcdot} - m_{p\lcdot} +  \, \proj_{\PP} \left(  2m_{p\lcdot} - m^1_{p\lcdot} - \frac{1}{L} \nabla J_\rho(M)_{p\lcdot} \right)
        	\;\Forall p \in \P$
        \STATE update $M_2$: $m^2_{\lcdot r}  \leftarrow  m^2_{\lcdot r} - m_{\lcdot r} +  \, \proj_{\PP_\le} \left(  2m_{\lcdot r} - m^2_{\lcdot r} - \frac{1}{L} \nabla J_\rho(M)_{\lcdot r}  \right)
            \; \Forall r\in \R^+$%\Forall I \in \II^+, \,\Forall r\in \R(I)
        \STATE update $M\gets \frac{1}{2} (M_1 + M_2)$  %\quad \text{ (matrix update) }
\ENDWHILE 
\end{algorithmic}
\normalsize
\end{algorithm}
%}
The positive parameter $L$ controls the gradient descent step. Experimentally, we set $L = \frac{1}{10} \norm{A}$ and estimate $\norm{A}$ using power-iteration.
The projector onto $\PP$ is computed in linear time~\cite{Condat_simplex}. The projection onto $\PP_\le$ is trivial.
Note that other splitting schemes are possible and have been tested (for instance, using non-negativity constraint on a third variable), but this combination was particularly efficient (faster convergence).
The main advantage of this algorithm
%, particularly over the Iterative soft-assign one,
is that it can be massively parallelized. %, in comparison to other approaches.

\section{Experiments}

%Part encoding
%We add spatial pyramid
%We add PCA to CoP
%Part selection ?

\subsection{Datasets}

\subsubsection{The Willow actions dataset \cite{delaitre10}}
is a dataset for action classification, which contains 911 images split into 7 classes of common human actions, namely \textit{interacting with a computer, photographing, playing music, riding cycle, riding horse, running, walking}.
There are at least 108 images per actions, with around 60 images used as training and the rest as testing images.
The dataset also offers bounding boxes, but we do not use them as we want to detect the relevant parts of images automatically.
% The dataset also offers bounding boxes fitted on humans performing the actions, but we do not use them as we want to detect the relevant parts of images automatically without any prior knowledge on the scenes.

\subsubsection{The MIT 67 scenes dataset \cite{Quattoni09}} is an indoor scene classification dataset, composed of 67 categories. 
These include stores (e.g. bakery, toy store), home (e.g. kitchen, bedroom), public spaces (e.g. library, subway), leisure (e.g. restaurant, concert hall), and work (e.g. hospital, TV studio).
Scenes may be characterized by their global layout (corridor), or by the objects they contain (bookshop).
Each category has around 80 images for training and 20 for testing.

%--------------------------------------------------------------

\subsection{Improved description and classification pipeline}\label{sec:desc}

We follow the general learning and classification pipeline of~\cite{Sicre15A}, however we also introduce significant improvements.
Such improvements makes sense in order to compete with recent works.
In summary, during part learning, $\card{\R}=1,000$ regions are extracted from each training image and used to learn the parts. During encoding, $\card{\R}$ regions are extracted from both training and test images, and all images are encoded based on the learned parts. Finally, a linear SVM is used to classify test images. For each stage, we briefly describe the choices made in~\cite{Sicre15A} and discuss our improvements.

\subsubsection{Initialization}\label{sec:init}
The initialization step is achieved as in~\cite{Sicre15A}.
All training positive regions are clustered and for each cluster an LDA classifier is computed over all regions of the cluster.
% These classifiers are further applied to training regions form both positive and negative sets.
Maximum responses to the classifiers are then selected per image and averaged over positive and negative sets to obtain two scores.
The ratio of these scores is used to select the top $\p$ clusters to build the initial part classifiers. Finally, an initial matching matrix $M$ is built by softmax on classifier responses. This scheme is followed for all optimization algorithms, even if a part model matrix is not explicitly formed during iterations.

\subsubsection{Extraction of image regions}  
Two strategies are investigated:
\begin{itemize}
\item \emph{Random regions} (`\rR'). As in~\cite{Sicre15A}, $|\R|$ regions are randomly sampled over the entire image. The position and scale of these regions are chosen uniformly at random, but regions are constrained to be square and have a size of at least 5\% of the image size.
\item \emph{Region proposals} (`\rP'). Following~\cite{Mettes15}, up to $|\R|$ regions are obtained based on selective search~\cite{vandeSande11}. If less than $\p$ regions are found, randomly sampled regions are added to complete the set.
\end{itemize}

\subsubsection{Region descriptors}
Again two strategies are investigated, based on  fully connected CNN or convolutional layers:
\begin{itemize}
\item \emph{Fully connected} (`FC'). As in~\cite{Sicre15A}, we use the output of the 7th layer of the CNN proposed by~\cite{jia14} on the rescaled region, resulting in a 4,096-dimensional vector. 
For the Willow dataset, we use the standard Caffe CNN architecture~\cite{jia14} trained on ImageNet. For MIT67, we use the hybrid network~\cite{zhou14p} trained on ImageNet and on the Places dataset. The descriptors are square-rooted and $\ell_2$-normalized.
\item \emph{Convolutional} (`C'). As an improvement, we use the last convolutional layer, after ReLU and max pooling, of the very deep VGG-VD19 CNN~\cite{Simonyan14c} trained on ImageNet. To obtain a region descriptor, we employ average pooling over the region followed by $\ell_2$-normalization, resulting in a 512-dimensional vector. Contrary to `FC', we do not need to rescale every region and feed it to the network; rather, the entire image is fed to the network only once, as in~\cite{HZRS14,Girs15}. Further, following~\cite{Tolias16}, pooling is carried out by an integral histogram. These two options enable orders of magnitude faster description extraction compared to `FC'. To ensure the feature map is large enough to sample $\card{\R}$ regions despite loss of resolution (by a factor of 32 in the case of VD-19), images are initially resized such that their maximum dimension is 768 pixels; this has been shown to be beneficial~\cite{zheng16good}.
\end{itemize}

% Note that the same description method is used to compute region descriptors within the CNN-on-parts descriptor, see Section~\ref{enc}.

% \red{
% We propose a different approach to extract region description (P+C), based on the convolutional layers of a deeper network and using a set of selected regions.
% In the previous works, most methods extract features from the fully connected layers, which require to rescale every region to feed it to the network.
% Recently, the work \cite{Tolias16} shows that when considering only convolutional layer (\ie discarding fully connected ones), rescaling can be avoided by subtracting the mean pixel to the input image.
% Then, a full sized image can be fed only once to the network and multiple region descriptors can be extracted directly from the feature maps, \ie the output of the network.
% This allows to extract thousands of regions descriptors very effectively, using integral histograms for instance.
% Furthermore, these descriptions are shorter (\eg 512 dimensional for the VD-19 network) than the ones based on fully connected layers (\eg 4096 dimensional for the Caffe network).
% We note however that the regions can not be very precise as the input image get its dimension reduced (by a factor of 32 in the case of VD-19).
% Therefore, we scale all input image to have a fixed maximum size and have a large enough feature map to extract a thousand regions from. 
% Furthermore, the work of \cite{zheng16good} showed that rescaling the entire database to the average image size yields improvements in terms of performance.
% }

\subsubsection{Encoding}%Ronan
Given an image, either training or testing, region descriptors are tested against the learned part model to generate a global image descriptor, which is then used by a SVM classifier. We use several alternative strategies:
\begin{itemize}
\item \emph{Bag-of-Parts} (`BoP') and \emph{Spatial Bag-of-Parts} (`SBoP'). According to BoP~\cite{Sicre15A}, for each part classifier, the maximum and average score is computed over all regions; the scores for all parts are then concatenated. Here we introduce SBoP, which adds weak spatial information to BoP by using Spatial Pyramids as~\cite{doersch13}. In this case, maximum scores are computed over the four cells of a $2\times2$ grid over the image and appended to the original BoP.
\item \emph{CNN-on-Parts} (`CoP') and \emph{PCA on CNN-on-Parts} (`PCoP'). According to CoP~\cite{Sicre15A}, the CNN descriptors corresponding to the maximum scoring region per part are concatenated to form the image descriptor. Here we also investigate PCoP, whereby centering and PCA is applied to CoP as in~\cite{Sicre15B}.
% ; this is shown to improve performance using either the convolutional or the fully connected layers~\cite{Sicre15B}.
\end{itemize}

% \subsubsection{Encoding}%Ronan
% % \label{enc}
% Concerning encoding, we propose to improve the method of \cite{Sicre15A}.
% Once parts are learned, a given image is described either by the Bag-of-parts (BoP) or CNN-on-parts (CoP).
% %BoP is obtained by first computing the per parts scores for each extracted region on an image.
% %Then, the descriptor is obtained by concatenating per parts the average and the maximum of the region scores over the image.
% %Moreover, CoP is the concatenation of the CNN descriptors of the maximum scoring region over the image for every part of the model.

% We propose to improve the BoP and CoP.
% SBoP adds weak spacial information to the BoP by using Spatial Pyramids as \cite{doersch13}.
% Maximum scores are computed over the four segments ($2\times2$) of the image and concatenated to the original BoP.
% PCoP applies centering and PCA to the CoP, as it is shown to improve performance of the fully connected layers \cite{Sicre15B}, as well as the convolutional layers.

% \red{
% In the second step, two different encoding strategies are considered: (i) Bag-of-parts (BoP) in which the maximum and average scores of each part classifier are aggregated per image and (ii) CNN-on-parts (CoP) in which the CNN descriptors of the per parts maximum scoring region per image are concatenated to form the image descriptor.
% Once images descriptors are computed, a linear SVM is applied for the final classification. 
% }

\subsubsection{Parameters of the learning algorithms} 
For the Iterative Soft-Assign (ISA) method, we use the same parameters as~\cite{Sicre15A}.
Concerning the GFB method, we perform 2k iterations of the projection, except for the MIT67 dataset with convolutional descriptor, where iterations are limited to 1k.
In all experiments performance remains stable after 1k iterations.
For the GFB method with $\rho \ne 0$, reffed to as $\text{GFB}_\rho$, we choose $\rho = 10^{-3} \norm{A}$ after experimental evaluation on the Willow dataset. We denote by just GFB the case where $\rho=0$.
%$\rho = 1/|\cal{R}|$.%, see Figure \ref{fig_GFB_rho} 

%--------------------------------------------------------------

%\begin{figure}[tb]
%\centering
%	\includegraphics[width=0.5\linewidth]{fig/GBF_rho.pdf}
%\caption{Evaluation of the $\rho$ parameter of $\text{GFB}_\rho$ on the Willow dataset} 
%\label{fig_GFB_rho}
%\end{figure}

%--------------------------------------------------------------

\subsection{Results}

%The original performance are presented in Table \ref{origin}. Note that the CoP encoding in the following is different and better performing (\~2\% better) than CoP (NE).

In the following, we are showing results for (i) fully connected layer descriptor on random regions (R+FC), which follows~\cite{Sicre15A}, and (ii) convolutional layer descriptor on region proposals (P+C), which often yields the best performance. We evaluate different learning algorithms on BoP and CoP encoding, and then investigate the new encoding strategies SBoP and PCoP as well as combinations for the ISA algorithm. On Willow we always measure mean Average Precision (mAP) while on MIT67 we calculate both mAP and classification accuracy (Acc).

We start by providing, in Table~\ref{tab_base}, a baseline corresponding to our description methods on the full image without any part learning. Comparing to subsequent results with part learning reveals that part-based methods always provide better description of the content of an image.

%--------------------------------------------------------------

\begin{table}[ht]
	\centering
	\caption{Baseline performance, without part learning. }
	\label{tab_base}
	\smallskip
	\begin{tabular}{|l|c||c|c||c|c|} \hline
	    \multirow{2}{*}{Method}       & \multirow{2}{*}{Measure} & \multicolumn{2}{c||}{Willow} & \multicolumn{2}{c|}{MIT67} \\ \cline{3-6}
	                                  &     & FC & C & FC & C \\ \hline\hline
	    \multirow{2}{*}{Full-image}   & Acc & -- & -- & 70.8 & 73.3 \\	
	                                  & mAP & 76.3   & 88.5   & 72.6 & 75.7 \\\hline
	\end{tabular}
\end{table}	

%--------------------------------------------------------------

\begin{figure}[htb]
\begin{center}
	\includegraphics[width=.8\linewidth]{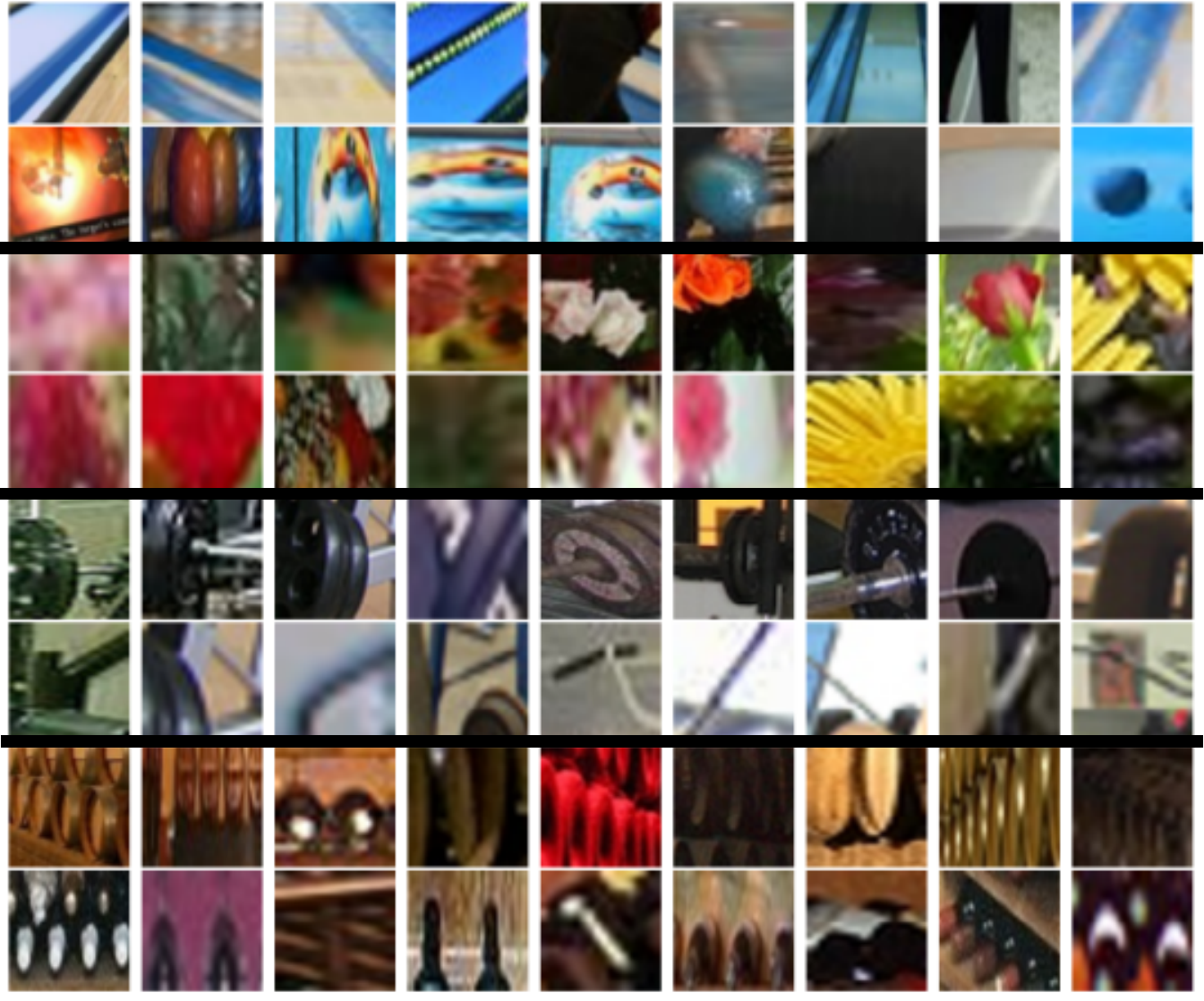}
	\smallskip
\caption{Top scoring parts for various images of bowling, florists, gym  and wine cellar} 
\end{center}
\label{fig_parts}
\end{figure}

We now focus on the part learning methods, which are evaluated in the context of action and scene classification in still images.
Figure \ref{fig_parts} shows some qualitative results of learned parts on MIT67.
%Table~\ref{tab_willRD} and \ref{tab_MITRD}
Then, Table~\ref{tab_P+C} shows the performance of ISA, IPFP, Hungarian, GFB, and $\text{GFB}_\rho$ on Willow and MIT67 datasets.
After some evaluation on both MIT67 and Willow, IPFP was not evaluated in further experiments since it performs on par with the Hungarian or worst, as previously explained in \ref{sec:IPFP}.
On the Willow dataset, we observe that $\text{GFB}_\rho$ $>$ GFB $>$ Hungarian and IPFP $>$ ISA.
However, on MIT67 the results are different and we have ISA $>$ Hungarian and GFB $>$ $\text{GFB}_\rho$.
%\color{red}
When using the improved P+C descriptor, we observe a similar trend for the BoP.
% ISA performs better than the Hungarian and GFB, which are better than $\text{GFB}_\rho$.
Nevertheless, note that all methods perform similarly when using the CoP encoding.

Based on this experimentation, we can draw two main conclusions.
% We conclude that soft-assign perform overall better than the other methods when using the Bag-of-Parts.
%
First, methods based on soft assignment (ISA, GFB) clearly outperform methods based on hard assignment. %, thus confirming previous results on this topic~\cite{Sicre15A}.
%\color{blue}
This result is also confirmed, in almost all cases, by the results of Table~\ref{tab_softhard}, where one iteration of the Hungarian algorithm is performed on the assignment matrix obtained after ISA (i.e. ISA+H).
%\normalcolor

\begin{table}[t]
	\centering
	\caption{Performance of various learning methods on Willow and MIT67.}
	\label{tab_P+C}
	\smallskip
	\begin{tabular}{|l|c||c|c|c|c|c|} \hline
		Method                            & Measure & ISA & IPFP & Hun & GFB & $\text{GFB}_\rho$ \\ \hline\hline
		Willow R+FC BoP                   &     & 76.6 & 79.0 & 78.9 & 79.7 & \textbf{80.6}  \\
		Willow P+C BoP                    & mAP & \textbf{89.2} & 86.3   & 88.3 & 88.2 & 87.5 \\ 
		Willow P+C CoP                    &     & 91.6 & 91.3   & 91.1 & \textbf{91.8} & \textbf{91.8} \\ \hline\hline	
		\multirow{2}{*}{MIT67 R+FC BoP}   & Acc & \textbf{76.6} & -- & 75.4 & 75.7 & 74.7 \\ 
		& mAP & \textbf{78.8} & -- & 78.0 & 77.6 & 76.3 \\ \hline
		\multirow{2}{*}{MIT67 P+C BoP}    & Acc & \textbf{75.1} & 70.7  & 72.8 & 70.9 & 70.9 \\
		& mAP & \textbf{76.7} & 72.6   & 75.1 & 73.5 & 73.1 \\ \hline
		\multirow{2}{*}{MIT67 P+C CoP}    & Acc & \textbf{80.0} & 79.2   & 79.8 & 79.2 & 79.3 \\
		& mAP & \textbf{80.2} & 79.7   & 79.9 & 79.5 & 79.7 \\ \hline
	\end{tabular}
\end{table}

%--------------------------------------------------------------

\begin{table}[ht]
	\centering
	\caption{Performance of Hard vs Soft assignment. 
		%\color{red}
		ISA+H refers to performing one iteration of the Hungarian algorithm on the solution obtained by ISA.}
	%\normalcolor}
	\label{tab_softhard}
	\smallskip
	\begin{tabular}{|l|c|c|c|} \hline
		Method                          & Measure & ISA & ISA+H \\ \hline
		Willow R+FC BoP                  &  mAP   & 76.6  & \textbf{76.9}\\
		Willow P+C BoP                 &  mAP   & \textbf{89.2}  & 88.1\\
		Willow P+C CoP                 &  mAP   & \textbf{91.6}  & 89.6\\
		MIT67 R+FC BoP                  &  mAP   & \textbf{78.8}  & 77.9  \\ \hline
	\end{tabular}
\end{table}	

%--------------------------------------------------------------

%\color{blue}
Second, while the GFB offers some significant practical advantage, 
when combined with quadratic regularization it is out-performed by Iterative Soft-Assign (except on the Willow dataset with BoP and CoP, first and third line in Table~\ref{tab_P+C}).
%\normalcolor
Our explanation is that it demonstrates that quadratic regularization is less appropriate than entropic regularization for this problem. 
Indeed, as illustrated in Section~\ref{sub:relax}, over-relaxing the objective function $J_\rho$ tends to yield a matrix with very similar rows, meaning that parts are described by the same regions, which is highly undesirable. While this problem also occurs when solving a soft-assignment with very large regularization, it does not happen when using ISA.

Another possible explanation of this difference in performance may lie in the fact that the Iterative Soft-Assign is stopped before convergence and does not satisfy the constraints $\PP$ imposed on rows, whereas those constraints are satisfied when using the GFB algorithm.
%\color{blue}
We conjecture that the constraint, \ie ``a part must occur in every positive image'' in the original problem definition [8], is too strong and may need to be relaxed.%and that we reach the limit of the problem definition."
%\normalcolor
Actually, as highlighted in the introduction (Sect.~\ref{intro}), the limitation of the separate optimization problem in comparison with the joint optimization is that a better optimization of the intermediate goal does not necessarily produce better final performance.

Focusing on the ISA method, the improved region description and encoding are evaluated, see Table \ref{tab_iEnc}.
Using region proposals along with convolutional layer descriptions shows a significant performance gain, especially on the Willow dataset.
We can see a consistent improvement for the SBoP and PCoP encoding as well and note that PCA yields more improvement on the descriptors based on fully connected layer than on the ones based on convolutional layers.
These improvements set a new state of the art on both datasets, obtaining 91.9\% mAP on Willow and 81.4\% mAP on MIT67. Table~\ref{tab_SoA} compares our best performance on MIT67 to a number of previous methods. Furthermore, we outperform the previous state of the art on Willow~\cite{Mettes15} with 81.7\% mAP.

\begin{table}[ht]
	\caption{Results on Willow and MIT67 datasets for the ISA method, with improved region descriptions P+C and improved encoding methods SBoP and PCoP. BoP+CoP and SBoP+PCoP refer to concatenated image descriptors.}
	\label{tab_iEnc}
	\smallskip
	\scalebox{0.9}{
	\centering
	\begin{tabular}{|l|c||c|c||c|c||c|c|} \hline
		Method                          & Measure & BoP & SBoP & CoP & PCoP & BoP+CoP & SBoP+PCoP \\ \hline\hline
        Willow R+FC                     & \multirow{2}{*}{mAP} & 76.6 & 78.7 & 81.6 & 82.4 & 81.9 & 82.6 \\
        Willow P+C                      &     & 89.2 & 90.1 & 91.6 & 91.7 & 91.8 & \textbf{91.9} \\ \hline\hline
        \multirow{2}{*}{MIT67 R+FC}     & Acc & 76.6 & 76.1 & 76.8 & 77.1 & 78.1 & 78.3 \\ 
                                        & mAP & 78.8 & 79.0 & 77.8 & 79.5 & 80.1 & 80.7 \\ \hline
        \multirow{2}{*}{MIT67 P+C}      & Acc & 75.1 & 76.1 & 80.0 & 80.5 & 81.1 & \textbf{81.4} \\ 
                                        & mAP & 76.7 & 76.7 & 80.2 & 81.0 & 81.0 & \textbf{81.2} \\ \hline
	\end{tabular}
}
\end{table}

%\begin{table}[h]
%	%\renewcommand{\arraystretch}{1.3}
%	\caption{Results on Willow and MIT67 datasets for the ISA method, with improved encoding methods SBoP and PCoP. }
%%	\label{origin}
%	\centering
%	
%	\begin{tabular}{|l|c|c|c|c|}		
%	\hline
%		\textbf{Method} & \textbf{Willow R+FC}  & \textbf{MIT67 R+FC Acc (mAP)} & \textbf{Willow P+C (mAP)}  & \textbf{MIT67 P+C Acc (mAP)} \\ %& \textbf{Boats (MAP)} \\
%		\hline
%		BoP & 76.6 & 78.8 & 76.6        & 75.1 (76.7) \\
%		SBoP & & &         & 76.1 (76.7)\\
%		\hline
%		CoP & 81.6 & 77.8 & 76.8        & 80.0 (80.2) \\
%		PCoP & & 79.5&         & 80.5 (81.0) \\
%		\hline
%		BoP \& CoP & 81.9 & 80.1 & 78.1        &  \\
%		SBoP \& PCoP & & &         & 81.4 (81.2) \\
%		\hline
%	\end{tabular}
%	
%\end{table}

%\begin{table}[t]
%	\label{Will_G2G5}
%	\centering
%	\caption{Willow ${M}^\star$ G5 for 1 or 2 k?. NE means not Enlarged}
%	\smallskip
%	\begin{tabular}{|l|c|c|c|c|c|c|c|} \hline
%		Method          & G5-2k 2e3   & G5-2k e3    & G5-2k 5e4 & G5-2k 2e4 & G5-2k e4  & & G2-2k\\
%		\hline
%		BoP               & 80.1    & 80.6     & 80.8    & 80.4  & 79.9  & & 79.7\\
%%		Hun BoP            & 78.6    & 79.1     & 79.2    & 79.0  & 79.0  & & 78.6 \\
%				\hline
%	Sparcity (\%)   & 3.69        & 15.08     & 31.05    & 47.73    & 54.82   & & \\
%		\hline
%	\end{tabular}
%
%\end{table}

%--------------------------------------------------------------

\begin{table}[ht]
\centering
\caption{Performance in terms of accuracy of existing part-based and non part-based methods on the MIT67 Scenes dataset.}
\smallskip
\label{tab_SoA}
\begin{tabular}{|l|c|c|} \hline
%     \multicolumn{1}{|c|}{\textbf{Method}} & \multicolumn{1}{|c|}{\textbf{MAP}}
% 	& \multicolumn{1}{|c}{\textbf{Dimensions}} \\ %
	\textbf{Methods} & Part-based &\textbf{MIT67}\\ \hline\hline
% 	bag-of-words & 0.345\\ %\cite{chatfield11}
% 	Fisher vectors & 0.550\\ %\cite{chatfield11}
% 	\midrule
%	Doersch \etal \cite{doersch13} & Yes & 66.9\\
    Zhou \etal \cite{zhou14p}& No &70.8\\
    Zuo \etal \cite{zuo2014learning} & Yes & 76.2\\
    Parizi \etal \cite{parizi2014automatic} & Yes &77.1\\
    Mettes \etal \cite{Mettes15}& Yes &77.4\\
    Sicre \etal \cite{Sicre15A} & Yes &78.1\\
    Zheng \etal \cite{zheng16good} & No & 78.4 \\
    Cimpoi \etal \cite{cimpoi15deep}& No & 81.0\\ \hline\hline            
    \textbf{Ours} & Yes & \textbf{81.4}\\ 
	%///BOW on parts combi on top ? & ??\\
    \hline
\end{tabular}
%}
\end{table}

\section{Conclusion}

To conclude, we have investigated in this work the problem of discovering parts for part-based image classification. 
%We have demonstrated that 
We have shown that this problem can be recast as a quadratic assignment problem with concave objective function to be minimized with non-convex constraints.
While being known to be a very difficult problem, several techniques have been proposed in the literature, either trying to find ``hard assignment" in a greedy fashion, or based on optimization of the relaxed problem, resulting in ``soft assignment''.
Several methods to address this task have been investigated and compared to the previous method of~\cite{Sicre15A} which achieves state of the art results.

We additionally proposed improvements on several stages of the classification pipeline, namely region extraction, region description and image encoding, using a recent very deep CNN architecture. This achieves a new state-of-the art performance on two different datasets. Furthermore, the new region description method is orders of magnitude faster, as this process was previously the bottleneck in~\cite{Sicre15A}.

%\color{red}
Our experiments show that, in the context of part-based image classification, soft assignment outperforms hard assignment. Moreover, entropic regularization is more appropriate than quadratic regularization, while the best overall performance is obtained when one constraint is not fully satisfied. While it is a common constraint to consider that a part must occur in every positive image, this interesting finding shows that this constraint may need to be relaxed.

Our reformulation and investigation of different optimization methods allow the exploration of the limits of the original problem, such as defined in~\cite{Sicre15A}. We believe this knowledge will help the community in the search for more appropriate models, potentially end-to-end trainable, using better network architectures.
%\normalcolor

\section*{Acknowledgment}
Part of this work was achieved in the context of the IDFRAud project ANR-14-CE28-0012.

\bibliography{egbib}

\end{document}